\documentclass[11pt]{article}

\usepackage[a4paper,margin=1in]{geometry}
\usepackage{graphicx}
\usepackage{booktabs}
\usepackage{amsmath,amssymb}
\usepackage{url}
\usepackage{multirow}
\usepackage[edges]{forest}
\usepackage{dirtree}
\usepackage{xurl}
\usepackage{fancyvrb}
\usepackage[
    style=nature,
    backend=biber,
    sorting=none,
    doi=true
]{biblatex}
\usepackage{float}
\usepackage{makecell}

\title{MotionScape: A Motion-Stratified UAV Video Benchmark for World Modeling and Future Video Generation}

\author{
Zile Guo$^{1,2,3}$, Zhan Chen$^{1,2}$, Xiaoxuan Liu$^{1,2,*}$, Enze Zhu$^{1,2,3}$,\\
Kan Wei$^{1,2,3}$, Yongkang Zou$^{1,2,3}$, Sijia Ma$^{4}$, Lei Wang$^{1,2}$ \\[2pt]
\parbox{0.9\textwidth}{\centering
$^{1}$Aerospace Information Research Institute, Chinese Academy of Sciences\\
$^{2}$Key Laboratory of Target Cognition and Application Technology,\\ Chinese Academy of Sciences\\
$^{3}$School of Electronic, Electrical and Communication Engineering,\\ University of Chinese Academy of Sciences\\
$^{4}$Xi'an Surveying \& Mapping Institute \\
$^{*}$liuxiaoxuan@aircas.ac.cn
}
}

\date{}

\begin{document}

\maketitle

\begin{abstract}

Unmanned aerial vehicles (UAVs) are increasingly crucial for low-altitude autonomy and complex environment understanding. World models enable UAVs to anticipate how future states may evolve under potential actions, providing predictive support for autonomous decision-making. For video world models, future video generation serves as a means of simulating future visual states, providing a direct basis for evaluating their predictive capability. However, existing UAV video resources typically focus on specific control signals or simulated environments. Standardized benchmarks for multi-condition UAV-view future video generation are still limited. To bridge this gap, we introduce MotionScape, a real-world UAV-view benchmark comprising 228 high-resolution video clips totaling 62,700 frames, with semantic annotations of weather and illumination conditions, scene environment, and camera-viewpoint motion. The clips are stratified into low-, medium-, and high-motion strata using optical-flow-based motion intensity. MotionScape supports Text2World, Image2World, and Video2World evaluation under a unified future-generation protocol. Technical validation with representative baseline models assesses performance across conditioning settings and motion strata, providing a unified platform for evaluating the future visual simulation capabilities of UAV world models.

\end{abstract}

\section*{Background \& Summary}

UAV platforms have broad application prospects in low-altitude autonomous flight, complex environment understanding, path planning, disaster search and rescue, and infrastructure inspection~\cite{article,kaufmann2023champion}. As UAV applications progress toward autonomous operation in open environments, UAV systems require the ability to understand current observations and anticipate future changes in the surrounding world~\cite{pmlr-v87-kaufmann18a}. World models are important for UAV agents because they go beyond understanding what is currently visible: they learn how a scene changes over time and use this knowledge to anticipate future states for planning and decision-making~\cite{NEURIPS2018_2de5d166,pmlr-v97-hafner19a,Hafner2025TrainingAI,Genie}. In particular, UAV agents need to anticipate how their observations will change as they move through the environment, which requires modeling the relationship between platform motion, viewpoint evolution, and scene appearance. A key capability is therefore to generate plausible future visual states under different semantic and visual conditions while maintaining consistent scene content and camera motion~\cite{he2025CameraCtrl}. Future visual generation thus provides a direct means of assessing the ability of video world models to simulate spatiotemporal evolution~\cite{pmlr-v267-kang25g,videoworldsimulators2024}. This is particularly challenging in real-world UAV videos, where rapid translation and rotation, large viewpoint changes, altitude variations, and frequent changes in visibility and object scale produce highly dynamic visual sequences~\cite{scirobotics.abg5810,He_2025_ICCV,rs18142323}. Models must preserve global scene structure, local visual details, and camera-motion adherence throughout the generated sequence.

\begin{table*}[t]
\centering

\caption{Comparison of MotionScape with UAV-view video datasets and benchmarks. N.R. denotes not reported.}
\resizebox{\textwidth}{!}{
\begin{tabular}{lllllll}
\toprule  
\textbf{Resource} & 
\textbf{Primary role} & 
\textbf{Conditioning} & 
\textbf{Resolution / FPS} & 
\textbf{Data source} \\
\midrule
Sekai~\cite{Sekai} & 
World exploration dataset & 
Text / Trajectory & 
720p / N.R. & 
Real / Game  \\

AirScape~\cite{AirScape} & 
Aerial world-model dataset & 
Visual / Motion & 
480 × 720 / N.R. & 
Real  \\

UZH-FPV~\cite{UZH-FPV} &
High-speed UAV dataset &
Visual / Pose &
640 × 480 / 30 &
Real & \\

OpenSafari~\cite{OpenSafari} & 
Camera-controlled UAV video dataset & 
Trajectory / Pose & 
720p / 24 & 
Real  \\

Mid-Air~\cite{Mid-Air} & 
Synthetic UAV simulation dataset & 
Sensors / Trajectory & 
1024 × 1024 / 35 & 
Synthetic  \\

\midrule
\textbf{MotionScape (ours)} & 
\textbf{Multi-condition evaluation benchmark} & 
\textbf{Text / Image / Video} & 
\textbf{$\geq$1080p / 29.97} & 
\textbf{Real}  \\
\bottomrule
\end{tabular}}
\label{tab:uav_dataset}
\end{table*}

As summarized in Table 1, several UAV-oriented datasets have recently been developed for video generation, navigation, simulation, and world-model research, yet publicly accessible benchmarks for UAV future video generation remain scarce~\cite{rs18142323,Sekai,AirScape,UZH-FPV,OpenSafari,Mid-Air}. The scarcity is more pronounced for multi-condition evaluation, where video generation and world models may operate under different forms of conditioning, including text, images, and preceding video frames~\cite{cosmos2.5}. Existing data resources generally focus on specific inputs or tasks, such as trajectory- or action-conditioned prediction, and do not provide a unified protocol for comparing text-, image-, and video-conditioned future generation. Moreover, they do not systematically assess the structural and temporal consistency challenges introduced by real-world UAV videos with highly dynamic viewpoints. These limitations motivate a benchmark for multi-condition evaluation of UAV future video generation, particularly in highly dynamic cases.

To address this issue, we introduce MotionScape, a real-world UAV-view video benchmark comprising 228 video clips for Text2World, Image2World, and Video2World evaluation under a unified protocol. Each video clip is accompanied by semantic annotations consisting of weather and illumination conditions, an environment description, and a caption describing scene content and camera-viewpoint motion. The clips are assigned to low-, medium-, and high-motion strata using an optical-flow-based score, enabling systematic comparisons across conditioning settings and motion strata.

MotionScape is constructed from publicly accessible UAV and first-person-view (FPV) videos through manual source selection, CLIP-assisted inspection, and human verification. We evaluate representative video generation and world models under the multi-condition future-generation protocol and compare their performance across conditioning settings and motion strata. The results show that MotionScape’s motion stratification captures increasing generation difficulty with stronger UAV-view motion, while clip-level descriptions provide usable semantic cues for future generation. Taken together, MotionScape enables systematic assessment of the future video generation capability of UAV world models across different motion conditions and conditioning settings.

\section*{Methods}

\subsection*{Data Collection}
\begin{figure*}
  \includegraphics[width=\textwidth]{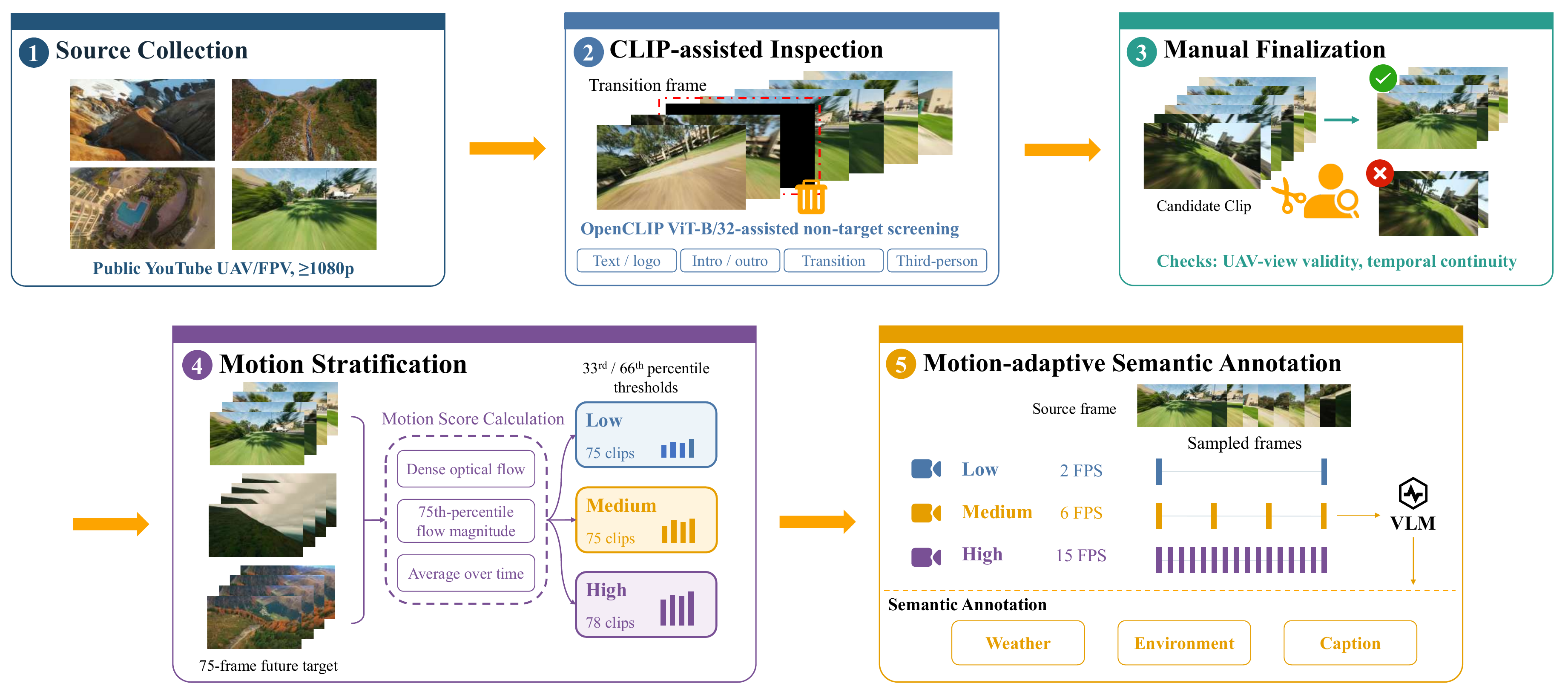}
  \caption{Construction and annotation pipeline of the MotionScape benchmark. Candidate clips retained after CLIP-assisted inspection are manually finalized with temporal-boundary adjustment, stratified into low-, medium-, and high-motion strata using optical-flow-based motion scores, and semantically annotated with motion-adaptive frame sampling. VLM denotes vision-language model.}
  \label{fig:pipeline}
\end{figure*}

The overall construction and annotation pipeline of MotionScape is summarized in Fig.~\ref{fig:pipeline}. The source videos used to construct MotionScape were collected from publicly accessible UAV and FPV aerial videos on YouTube (\url{https://youtube.com}). We first used “FPV” as the primary search keyword, manually browsed relevant video collections and playlists, and selected real-world UAV videos captured from a first-person perspective as candidates. For videos included in the candidate pool, we recorded their original YouTube URLs to ensure data-source traceability. Candidate videos with resolutions below 1080p were excluded before downloading, and the downloaded files were subsequently verified to meet the same resolution requirement. The candidate videos cover various scenes, including urban areas, rural regions, mountains, coastlines, forests, roads, and building clusters, under different weather and illumination conditions. For the downloaded source videos, we recorded basic file-level information, including resolution, frame rate, and duration, for subsequent CLIP-assisted inspection.

\subsection*{CLIP-assisted Quality Inspection}
After manual pre-screening of the candidate videos, we used CLIP~\cite{CLIP} to assist in locating residual non-target segments within the candidate videos. Specifically, we used the OpenCLIP ViT-B/32 model~\cite{OpenCLIPViT-B/32} with OpenAI pretrained weights to sample frames from the candidate videos at 5 FPS and match the sampled frames against a set of negative text prompts. At this stage, we used only negative prompts with CLIP to identify residual non-target segments.

The negative text prompts were grouped into four categories: subtitles or text overlays, title or logo intro/outro screens, transition effects, and third-person drone shots. For each sampled frame, we computed the similarity between its image feature and the text features of the negative prompts in each category, and used the maximum similarity within each category as its abnormality score. If the score of a category exceeded its corresponding threshold, the frame was flagged as a candidate abnormal frame of that category. For the transition effect category, we combined CLIP matching with a detector based on frame brightness changes. A frame was flagged as a transition frame only when it was detected by the brightness-based detector and its transition-category CLIP abnormality score exceeded the corresponding similarity threshold, reducing false detections.

We then merged temporally adjacent abnormal sampled frames into candidate abnormal segments and removed isolated single-frame abnormal intervals to suppress incidental false positives. The script outputs the candidate abnormal time intervals for each video, their corresponding abnormal categories, and the remaining valid time intervals. The retained valid temporal intervals were subsequently used for candidate-clip sampling and manual finalization.

\subsection*{Candidate Video Clip Sampling and Finalization}

Based on the valid temporal intervals retained after CLIP-assisted inspection, candidate windows of approximately 10--15 seconds were randomly sampled from these valid intervals, with up to five candidates retained for each source video. Candidate windows from the same source video were constrained to be non-overlapping and separated by at least 2 seconds, and a fixed random seed of 20260505 was used for reproducibility. These windows were used for manual review and temporal-boundary adjustment. During manual finalization, temporal inconsistencies near clip boundaries were corrected by adjusting the temporal range whenever possible, whereas candidates that could not be repaired were discarded. For each finalized candidate, a fixed-length benchmark clip of 275 consecutive frames was then extracted from the finalized start timestamp after temporal resampling to 29.97 FPS, corresponding to approximately 9.2 seconds. Each benchmark clip was partitioned into a 200-frame observed prefix and a continuous 75-frame future target.

\subsection*{Motion Intensity Estimation and Stratification}

To characterize the influence of UAV dynamics on future video generation, we compute an optical-flow-based motion-intensity score for each clip following the procedure illustrated in Fig.~\ref{fig:pipeline}. Since MotionScape primarily consists of onboard FPV footage, the camera viewpoint moves together with the UAV, and changes in UAV motion are reflected in the evolution of the camera viewpoint. We therefore use image-space optical-flow displacement as a practical proxy for UAV-view visual dynamics~\cite{Wulff_2017_CVPR}. Because the motion stratification is intended to characterize the dynamics of the future target, the motion score is computed only over the 75-frame future target segment (frames 201--275).

For motion-score computation, each future target segment is temporally downsampled by selecting one frame every three frames, resulting in an effective frame rate of approximately 9.99 FPS. The sampled frames are resized to 854×480, and dense optical flow is computed between each pair of adjacent sampled frames using the Farnebäck method~\cite{farneback2003}. Let $I_t$ and $I_{t+1}$ denote two adjacent sampled frames, and let $\mathbf{F}_t(x,y)$ denote the corresponding dense optical-flow vector at spatial location $(x,y)$. Given a sequence of T sampled frames, the clip-level motion score is defined as
\begin{equation}
s_{\mathrm{clip}}
=
\frac{1}{T-1}
\sum_{t=1}^{T-1}
P_{75}^{x,y}
\left(
\left\|\mathbf{F}_t(x,y)\right\|_2
\right),
\end{equation}
where $\left\|\mathbf{F}_t(x,y)\right\|_2$ denotes the optical-flow magnitude at location $(x,y)$, and $P_{75}^{x,y}$ denotes the 75th percentile over all spatial locations. For each adjacent frame pair, the spatial 75th percentile captures prominent visual displacement while reducing sensitivity to isolated extreme values. Averaging these pair-level statistics over time produces a single motion-intensity score for each clip. Based on the distribution of clip-level motion scores across the benchmark, the 33rd and 66th percentiles are used as thresholds to divide the clips into low-, medium-, and high-motion strata. 

\subsection*{Semantic Annotation with Camera-Motion Descriptions}
At this stage, we generate semantic annotations for each video clip, as outlined in Fig.~\ref{fig:pipeline}, which are used as part of the textual conditioning in the benchmark tasks. We use GPT-4o mini~\cite{openai2024gpt4omini} to annotate each retained video clip. To help the annotation model infer how the camera-viewpoint motion evolves into the future target segment, the visual input includes the five terminal observed frames, corresponding to frames 196--200, together with sampled frames from the future target segment, corresponding to frames 201--275. Frames 196--200 are provided solely as temporal context, while the annotation model is instructed to describe only the future target segment spanning frames 201--275.

Because the benchmark clips exhibit substantial variation in motion intensity, a fixed temporal sampling stride may not adequately balance temporal coverage and visual redundancy. We therefore adopt a motion-adaptive frame-sampling strategy based on the motion stratum of each clip. Specifically, frames 196--200 are always retained as temporal context, while the future target frames 201--275 are sampled using temporal strides of 15, 5, and 2 for the low-, medium-, and high-motion strata, respectively.

Each annotation record contains three fields: weather, environment, and caption. The weather field describes the weather or illumination condition in the described interval, including common cases such as sunny, overcast, rainy, fog/haze, and night scenes. Unknown is used when the weather or illumination condition cannot be reliably determined from the visual content. The environment field provides a short free-form description (up to five words) of the main scene attributes. This free-form format accommodates the compositional nature of real-world UAV scenes. The caption field consists of a 2--3-sentence natural-language description covering the principal scene content and overall camera-viewpoint motion from frame 201 through frame 275. For benchmark conditioning, the weather, environment, and caption fields are combined to form the clip-level description.

\subsection*{Benchmark Tasks}

\begin{figure*}[t]
    \centering
    \includegraphics[width=\textwidth]{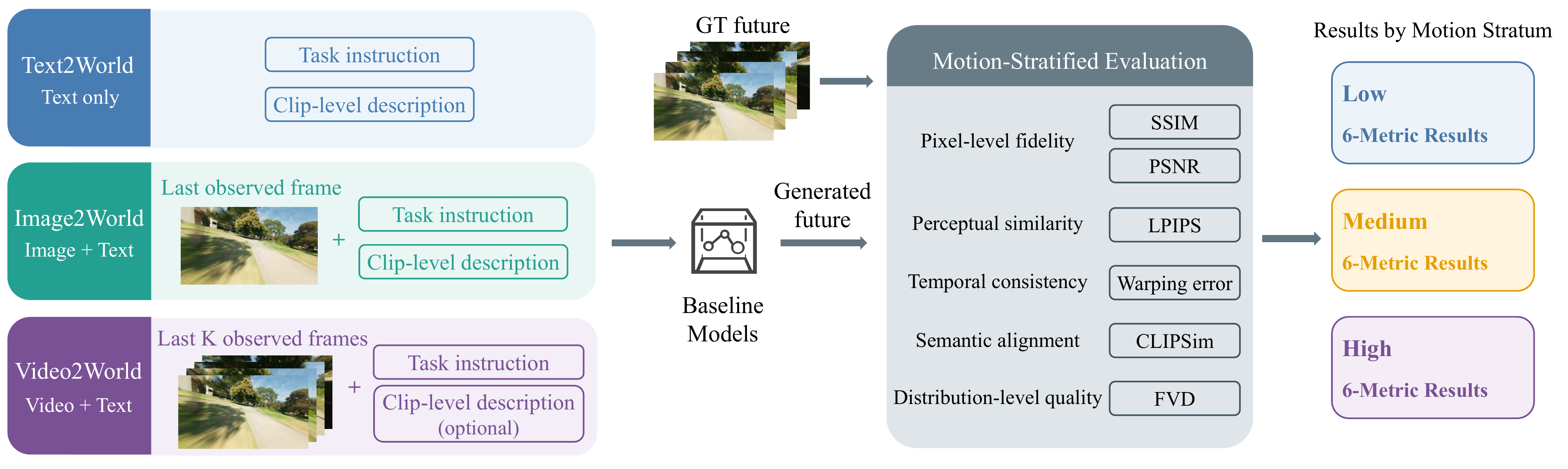}
    \caption{Overview of the MotionScape benchmark protocol. The multi-condition evaluation workflow covers Text2World, Image2World, and Video2World under a unified future-generation protocol with motion-stratified reporting. K denotes the model-specific number of terminal observed frames.}
    \label{fig:benchmark_overview}
\end{figure*}

An overview of the benchmark protocol is provided in Fig.~\ref{fig:benchmark_overview}. MotionScape evaluates future video generation under three conditioning settings, referred to as Text2World, Image2World, and Video2World following the terminology of Cosmos-Predict2.5~\cite{cosmos2.5}. Here, “World” denotes the future visual state represented in video form. Each benchmark sample contains a 200-frame observed prefix followed by a continuous 75-frame future target. For visually conditioned settings, conditioning frames are selected from the observed prefix. Text2World uses textual conditioning only. Image2World uses the final observed frame (the 200th frame) together with textual conditioning. Video2World uses a terminal sequence of observed frames together with textual conditioning, with the number of observed frames determined by each model’s native inference interface.

To standardize textual conditioning, each conditioning setting uses a fixed task-level instruction shared across all clips that defines the generation task and specifies basic requirements for preserving scene content and maintaining coherent viewpoint evolution. For Text2World and Image2World, the corresponding clip-level description, consisting of the weather, environment, and caption fields, is appended to this task-level instruction under “Future target description:”. For Video2World, we consider two textual settings under the same visual conditioning context: a task-level setting that uses only the fixed continuation instruction and a clip-level setting that additionally incorporates the corresponding clip-level description.

\subsection*{Quality Control}

To ensure the reliability and usability of MotionScape, we apply a quality-control procedure that combines manual review with automated integrity and consistency checks across all 228 clips. The procedure addresses three aspects: clip validity and temporal continuity, annotation reliability, and file integrity and cross-record consistency. 

All 228 final clips were manually verified after CLIP-assisted inspection and boundary adjustment, confirming valid UAV-view or first-person aerial content, temporal continuity, and the absence of residual non-target material.

We also performed targeted quality control on the semantic annotations. During preliminary model inference using the first-version captions, visible UAV bodies appeared in some generated videos. To reduce the possibility that annotation prompt wording contributed to this phenomenon, we revised the original prompt to require camera-centric descriptions and to avoid explicit carrier-related expressions, including “UAV-view”, “UAV”, and “drone”. The \texttt{caption} fields of all clips were subsequently regenerated using the revised prompt. To further assess the reliability of the semantic annotations, we conducted a stratified random check by sampling 10 clips from each motion stratum (30 clips in total). For each sampled clip, the environment description and scene-content description in the caption were manually compared with the corresponding visual content in the future target segment, while the described camera-viewpoint motion was compared with the overall observed motion trend. In the initial annotations, 27 clips were labeled as \texttt{Unknown}, primarily in cases with limited or ambiguous visual cues for weather identification. After manual inspection of these clips, only 3 clips remained labeled as \texttt{Unknown}, while the remaining 24 records were corrected with valid weather or illumination labels.

Automated integrity checks were performed on the finalized benchmark samples during construction. Each sample was verified to contain 275 consecutive frames, with the first 200 frames forming the observed prefix and the remaining 75 frames forming the future target. The released source metadata, annotation files, and motion-stratification assignments were checked for consistent identifiers and one-to-one correspondence. No missing, duplicated, or unmatched records were found.

\section*{Data Records}

\subsection*{Repository Structure}

\begin{figure}[H]
    \centering
    \includegraphics[width=\textwidth]{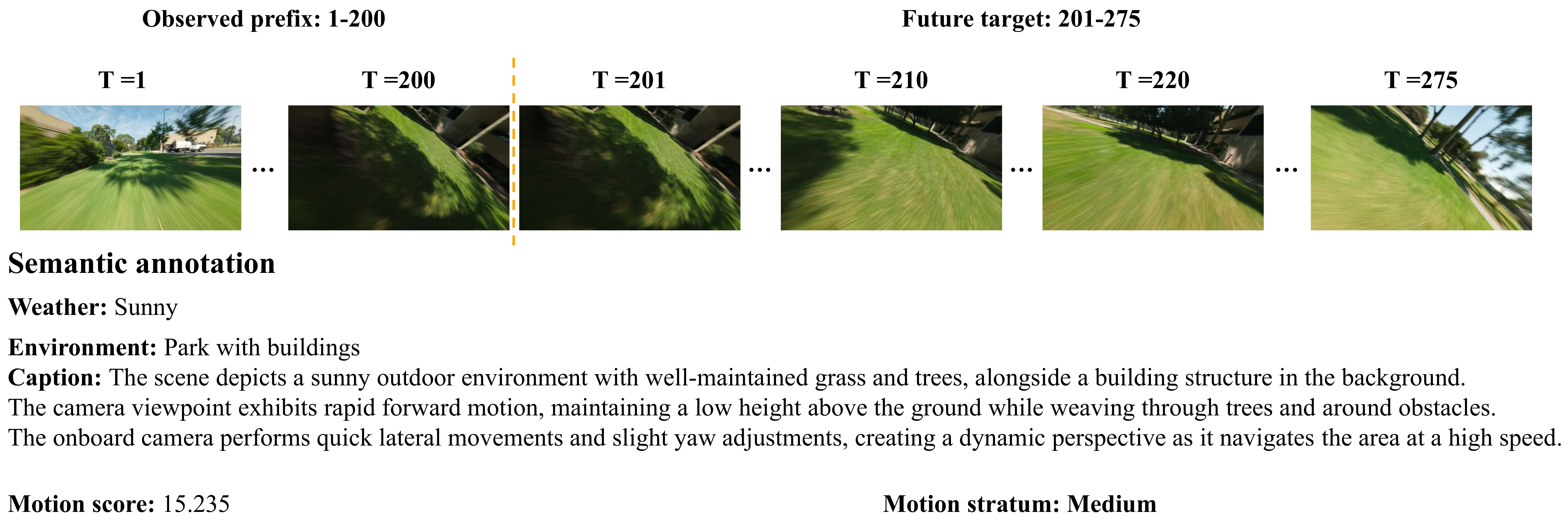}
    \caption{Representative MotionScape benchmark sample. Selected frames from the 200-frame observed prefix and the 75-frame future target are shown together with the corresponding semantic annotation and optical-flow-based motion score and stratum.}
    \label{fig:sample}
\end{figure}

The released MotionScape benchmark is organized into source video metadata, clip-level annotations, and motion-stratification assignments. Since MotionScape was constructed from publicly available YouTube UAV and FPV videos, source video metadata are provided to ensure data traceability and support reproducible benchmark reconstruction. The original audiovisual content is not redistributed as part of the release. The \texttt{source\_manifest.json} file records the original video URLs, the finalized start timestamp of each benchmark sample, and source video properties including resolution, frame rate, and video encoding information. Together with the released reconstruction code, these metadata enable reproducible reconstruction of MotionScape benchmark samples from the original video sources. The reconstruction process produces a 200-frame observed-prefix video, a 75-frame future-target video, the complete 275-frame video, and the corresponding 275-frame image sequence for each benchmark sample. Each clip is accompanied by a JSON annotation containing three primary fields: \texttt{weather}, \texttt{environment}, and \texttt{caption}. Motion-stratification assignments for all benchmark clips are recorded in \texttt{dynamicity\_buckets.json}.  A representative benchmark clip, together with its corresponding semantic annotation and motion-stratification information, is shown in Fig.~\ref{fig:sample}.

\subsection*{File Formats and Indexing}

Metadata and annotations are provided in JSON format. Each benchmark clip is identified by
\texttt{<source\_video\_name>\_\_segment\_<segment\_id>},
where \texttt{<source\_video\_name>} identifies the original source video and \texttt{<segment\_id>} identifies the extracted segment within that video. Segment indices are zero-based and represented using three-digit zero padding, beginning with \texttt{segment\_000}.

The identifier is also used across all released metadata components, including \texttt{source\_\allowbreak{}manifest.json}, clip-level annotation files, and \texttt{dynamicity\_buckets.json}. When benchmark samples are reconstructed locally using the provided reconstruction code, the generated video clips and extracted frame directories retain the same identifier, enabling direct correspondence between reconstructed samples and released metadata.

\section*{Data Overview}

\begin{figure}[H]
  \includegraphics[width=\textwidth]{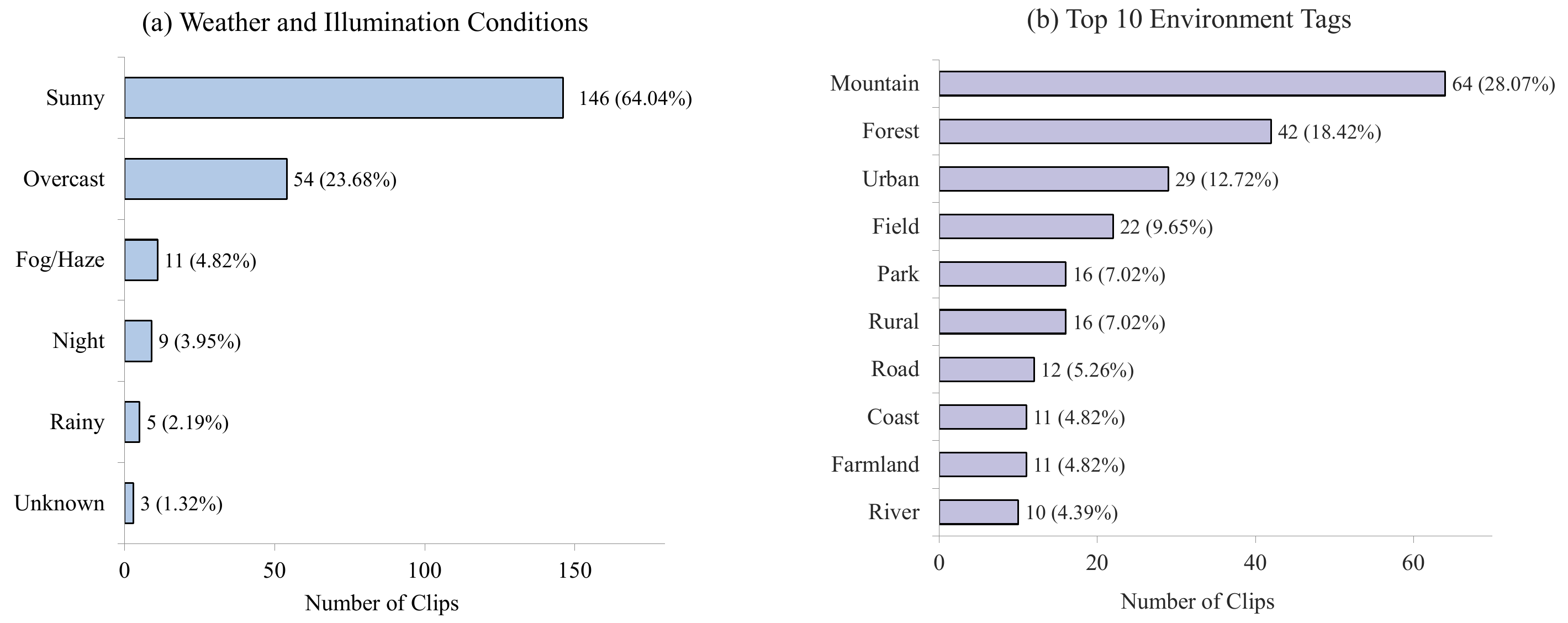}
  \caption{Environmental composition of the MotionScape benchmark. (a) Distribution of weather and illumination conditions across the 228 clips. (b) Clip-level prevalence of the ten most frequent normalized environment tags. A clip may be associated with multiple environment tags; therefore, the percentages in (b) do not sum to 100\%.}
  \label{fig:distribution}
\end{figure}

MotionScape contains 228 clips and 62,700 frames in total. Each clip consists of 275 frames standardized to 29.97 FPS, corresponding to approximately 9.2 seconds, and the complete benchmark has a total duration of approximately 34.9 minutes. The observed prefixes and future target segments contain 45,600 and 17,100 frames, respectively. The benchmark clips were derived from 67 distinct source videos, with an average of 3.4 clips selected from each source. Following the optical-flow-based motion stratification defined above, the benchmark includes 75 low-motion, 75 medium-motion, and 78 high-motion clips. The resulting motion-intensity statistics are summarized in Table~\ref{tab:flow_bucket_statistics}, showing clear differences in motion scores among the three strata. The captions contain 73.89 words on average, with a standard deviation of 8.21 words. The distributions of weather and illumination conditions and normalized environment tags are shown in Fig.~\ref{fig:distribution}. For visualization in Fig.~\ref{fig:distribution}(b), synonymous or closely related free-form environment descriptions were merged into normalized clip-level tags. This normalization was used solely for statistical visualization and did not modify the released annotations.

\begin{table}[t]
\centering
\small
\caption{Optical-flow statistics for the three motion strata. Mean, Std., Min., and Max. denote the mean, standard deviation, minimum, and maximum motion-intensity scores within each stratum.}
\label{tab:flow_bucket_statistics}
\setlength{\tabcolsep}{9pt}
\renewcommand{\arraystretch}{1.08}
\begin{tabular}{@{}lccccc@{}}
\toprule
\textbf{Motion Stratum} & 
\textbf{\# Clips} & 
\textbf{Mean} & 
\textbf{Std.} & 
\textbf{Min.} & 
\textbf{Max.} \\
\midrule
Low    & 75 & 4.087  & 3.081 & 0.001  & 9.262  \\
Medium & 75 & 12.727 & 2.052 & 9.386  & 15.983 \\
High   & 78 & 20.163 & 3.377 & 16.012 & 35.607 \\
\bottomrule
\end{tabular}
\end{table}

\section*{Technical Validation}

\subsection*{Motion-Stratified Evaluation Design}
All models and conditioning settings are evaluated using the same fixed low-, medium-, and high-motion clip assignments. This protocol supports direct comparison of model sensitivity to increasing UAV-view dynamics.

\subsection*{Evaluation Metrics}
We evaluate generated videos in terms of pixel-level fidelity, perceptual similarity, semantic alignment, temporal consistency, and distribution-level video quality. Peak Signal-to-Noise Ratio (PSNR) and Structural Similarity Index (SSIM)~\cite{SSIM} measure pixel-level similarity, Learned Perceptual Image Patch Similarity (LPIPS)~\cite{LPIPS} evaluates perceptual discrepancy, CLIP Similarity (CLIPSim)~\cite{CLIP} measures semantic alignment between the generated video and the corresponding caption field, and warping error evaluates temporal coherence~\cite{Lai_2018_ECCV}. We additionally report Fr\'echet Video Distance (FVD)~\cite{Unterthiner2019FVDAN} based on Inflated 3D ConvNet (I3D)~\cite{I3D} video features to characterize distribution-level similarity between generated and reference videos.

Because the evaluated models may produce videos at different native frame rates, generated frames are temporally aligned with the 75-frame reference target according to their timestamps before computing reference-based frame metrics. Each generated frame within the common target duration is paired with the temporally nearest reference frame, without temporal interpolation of the generated outputs. Before computing PSNR, SSIM, LPIPS, and warping error, generated and reference frames are resized to $1280 \times 704$. LPIPS uses the official implementation with an AlexNet backbone. CLIPSim uses \path{openai/clip-vit-base-patch32}, with the \texttt{caption} field of each clip used as the textual reference. CLIP similarity is computed for every generated frame within the evaluated target duration, and the frame-level similarities are averaged to obtain a clip-level CLIPSim score. Warping error is computed by warping each generated frame using Farneb\"ack optical flow~\cite{farneback2003} estimated from the corresponding consecutive reference frames and measuring the mean absolute error against the subsequent generated frame over valid regions. FVD is computed using the official \path{deepmind/i3d-kinetics-400/1} I3D network. Within each motion stratum, features from all generated and reference videos are aggregated separately, and a single Fr\'echet distance is computed between the two feature distributions. FVD is therefore reported at the motion-stratum level, whereas the remaining metrics are averaged from clip-level scores.

Several metrics require task-specific interpretation. In Text2World, PSNR and SSIM serve only as reference-based measurements because, without a visual prior, the same textual description may correspond to multiple visually plausible futures. In task-level Video2World, the same applies to CLIPSim because the corresponding caption is not provided as model input. Warping error is computed at each model's native output frame rate and is therefore interpreted only within the same model.

\subsection*{Baseline Models and Inference Settings}

We evaluate representative video generation and world models using their official checkpoints and native inference configurations, without fine-tuning. The model-specific temporal inference settings are summarized in Table~\ref{tab:temporal_settings}. In particular, the sampled conditioning frames for MAGI-1-24B were encoded at 16 FPS to match its native input configuration. For each model, the closest natively supported output length covering the benchmark future-target duration was selected. All inference experiments were conducted using four NVIDIA RTX PRO 6000 GPUs.

\begin{table}[H]
\centering
\caption{Temporal inference settings of the baseline models on MotionScape. Cond. denotes conditioning. Slash-separated entries follow Text2World / Image2World / Video2World order where applicable.}
\label{tab:temporal_settings}
\resizebox{\linewidth}{!}{%
\begin{tabular}{lcccc}
\toprule
Model
& \# Cond. Frames
& Cond. Sampling
& Native FPS
& \# Output Frames \\
\midrule
Cosmos-Predict2.5-2B
& 0 / 1 / 5
& -- / Final frame / Consecutive
& 30
& 93 / 92 / 88 \\
CogVideoX1.5-5B
& 0 / 1
& -- / Final frame
& 16
& 41 \\
Wan2.2-A14B
& 0 / 1
& -- / Final frame
& 16
& 40 \\
LongCat-Video
& 13
& Every 2 frames
& 15
& 40 \\
MAGI-1-24B
& 32
& Every 2 frames
& 16
& 40 \\
\bottomrule
\end{tabular}
}
\end{table}

\subsection*{Multi-condition Baseline Evaluation}

\begin{table}[t]
\centering
\caption{Multi-condition future video generation results on MotionScape across different motion strata. PSNR and SSIM for Text2World and CLIPSim for task-level Video2World are reported for reference only and are not used for direct comparison across conditioning settings.}
\label{tab:video_continuation_results}
\resizebox{\linewidth}{!}{
\begin{tabular}{lllcccccc}
\toprule
Mode & Model & Motion Stratum & PSNR $\uparrow$ & SSIM $\uparrow$ & LPIPS $\downarrow$ & Warping Error $\downarrow$ & FVD $\downarrow$ & CLIPSim $\uparrow$ \\
\midrule
\multirow{9}{*}[-1.8mm]{Text2World}
& \multirow{3}{*}{Cosmos-Predict2.5-2B~\cite{cosmos2.5}}
& Low     & 11.485  & 0.604 & 0.631 &  0.012 & 1367.287 & 0.317  \\
& & Medium & 10.712  & 0.485 & 0.724  & 0.025 & 1554.504 & 0.310  \\
& & High   & 10.660  & 0.431 & 0.756  & 0.032  & 1336.015 & 0.317  \\
\cmidrule(lr){2-9}
& \multirow{3}{*}{Wan2.2-T2V-A14B~\cite{Wang2025Wan}}	
& Low    & 10.322  & 0.478 &  0.687 & 0.048 & 1226.221 & 0.311 \\
& & Medium & 9.424  & 0.394 & 0.737 & 0.073 & 1238.102 & 0.308 \\
& & High   & 9.801  & 0.344 & 0.750 & 0.086 & 1360.666 & 0.317 \\
\cmidrule(lr){2-9}
& \multirow{3}{*}{CogVideoX1.5-5B~\cite{CogVideoX}}	
& Low    & 10.379  & 0.556 & 0.648 & 0.028 & 1467.129 & 0.317 \\
& & Medium & 9.525  & 0.432 & 0.738 & 0.047 & 1634.860 & 0.310 \\
& & High   & 9.746  & 0.376 & 0.759 & 0.061 & 1289.957 & 0.317 \\
\midrule
\multirow{9}{*}[-1.8mm]{Image2World}
& \multirow{3}{*}{Cosmos-Predict2.5-2B}
& Low    & 17.828  & 0.735 & 0.392  & 0.013  & 621.858 & 0.305 \\
& & Medium & 14.245 & 0.555 & 0.556  & 0.030  & 783.691 & 0.297 \\
& & High   & 13.022  & 0.445 & 0.631  & 0.044  & 766.657 & 0.305 \\
\cmidrule(lr){2-9}
& \multirow{3}{*}{Wan2.2-I2V-A14B}	
& Low     & 17.822 & 0.734 & 0.371 & 0.027 & 446.718 & 0.310 \\
& & Medium & 14.094 & 0.538 & 0.534 & 0.056 & 547.127 & 0.300 \\
& & High   & 12.549 & 0.420 & 0.604 & 0.077 & 483.050 & 0.305  \\
\cmidrule(lr){2-9}
& \multirow{3}{*}{CogVideoX1.5-5B-I2V}
& Low   & 16.971 & 0.699 & 0.395 & 0.021 & 772.220 & 0.309 \\
& & Medium & 13.531 & 0.522 & 0.566 & 0.044 & 1029.109 & 0.297 \\
& & High   & 12.144 & 0.420 & 0.639 & 0.061 & 888.128 & 0.299  \\
\midrule
\multirow{12}{*}[-2.4mm]{Video2World} 
& \multirow{3}{*}{\makecell[l]{Cosmos-Predict2.5-2B\\Task-level}}
& Low    & 18.873 & 0.764 & 0.351 & 0.018 & 468.377 & 0.293  \\
& & Medium & 14.994 & 0.573 & 0.522 & 0.039 & 652.729 & 0.278 \\
& & High   & 13.215 & 0.443 & 0.616 & 0.058 & 705.510 & 0.274 \\
\cmidrule(lr){2-9}
& \multirow{3}{*}{\makecell[l]{Cosmos-Predict2.5-2B\\Clip-level}}
& Low    & 19.024 & 0.760 & 0.360 & 0.016 & 449.789 & 0.305 \\
& & Medium & 14.998 & 0.576 & 0.527 & 0.035 & 596.103 & 0.299 \\
& & High   & 13.539 & 0.463 & 0.614 & 0.050 & 557.279 & 0.308 \\
\cmidrule(lr){2-9}
& \multirow{3}{*}{LongCat-Video~\cite{Cai2025LongCatVideo}}
& Low    & 20.183 & 0.768 & 0.325 & 0.027 & 336.065 & 0.308 \\
& & Medium & 15.088 & 0.558 & 0.505 & 0.061 & 517.162 & 0.296 \\
& & High   & 13.521 & 0.439 & 0.590 & 0.083 & 424.139 & 0.302 \\
\cmidrule(lr){2-9}
& \multirow{3}{*}{MAGI-1-24B~\cite{Teng2025MAGI1AV}}
& Low    & 19.837 & 0.763 & 0.352 & 0.027 & 411.466 & 0.303 \\
& & Medium & 14.684 & 0.559 & 0.540 & 0.057 & 575.835 & 0.289 \\
& & High   & 12.937 & 0.426 & 0.637 & 0.079 & 581.897 & 0.295 \\
\bottomrule
\end{tabular}
}
\end{table}

Table~\ref{tab:video_continuation_results} reports baseline performance across the three conditioning settings and motion strata. A clear motion-dependent degradation is observed under the visually conditioned Image2World and Video2World settings. From the low- to high-motion strata, PSNR decreases by 27.0--34.8\% and SSIM by 39.1--44.2\%, while LPIPS increases by 61.0--81.5\% and warping error by 185.2--238.5\%. FVD also generally deteriorates at higher motion strata, although it is not strictly monotonic across the intermediate stratum. This consistent cross-model trend indicates that the proposed motion stratification captures a meaningful increase in future-generation difficulty. The trend is less pronounced for Text2World, where the absence of visual context makes the target future substantially under-constrained; its PSNR and SSIM are therefore retained primarily as reference measurements.

In Text2World, Wan2.2-T2V-A14B achieves the lowest FVD in the low- and medium-motion strata (1226.221 and 1238.102), while CogVideoX1.5-5B achieves the lowest FVD in the high-motion stratum (1289.957). In Image2World, Cosmos-Predict2.5-2B achieves the highest PSNR (17.828, 14.245, and 13.022) and SSIM (0.735, 0.555, and 0.445) across the three motion strata, whereas Wan2.2-I2V-A14B obtains the lowest LPIPS (0.371, 0.534, and 0.604) and FVD (446.718, 547.127, and 483.050). In Video2World, LongCat-Video obtains the lowest LPIPS (0.325, 0.505, and 0.590) and FVD (336.065, 517.162, and 424.139) across all motion strata, while the clip-level Cosmos-Predict2.5-2B achieves the highest SSIM in the medium- and high-motion strata (0.576 and 0.463). 

CLIPSim exhibits a distinct pattern across conditioning settings. For Cosmos-Predict2.5-2B, Text2World achieves the highest CLIPSim across all three motion strata (0.317, 0.310, and 0.317), despite the absence of visual context. This suggests that text-only generation allows greater flexibility in following semantic descriptions, whereas visually conditioned generation requires the model to simultaneously preserve the observed scene and satisfy textual constraints.

Importantly, the task-level and clip-level variants of Cosmos-Predict2.5-2B provide a controlled comparison of textual conditioning. With the same visual context, introducing semantic annotations reduces warping error and FVD across all motion strata, with larger FVD improvements under medium and high motion. In contrast, changes in frame-level metrics are relatively modest. This suggests that semantic annotations provide complementary conditioning information beyond the observed visual context, especially under medium and high motion.

\subsection*{Failure Case Analysis}

\begin{figure}[H]
  \includegraphics[width=\linewidth]{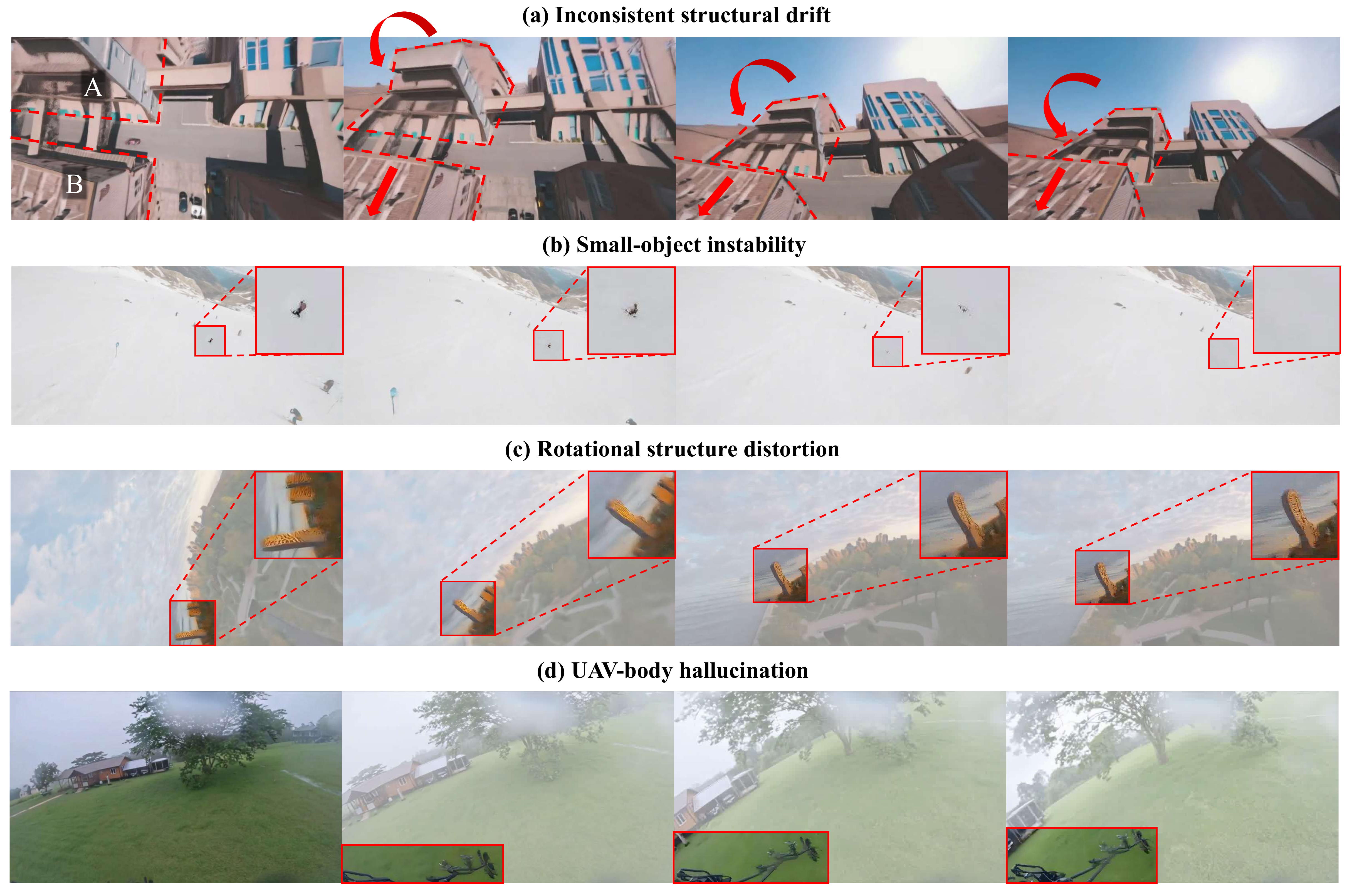}
  \caption{Representative failure cases observed during MotionScape evaluation. Red boxes highlight the affected regions, and arrows in (a) indicate viewpoint evolution.}
  \label{fig:FailureCase}
\end{figure}

Fig.~\ref{fig:FailureCase} presents representative cases illustrating the challenging generation behaviors exposed by MotionScape. In Fig.~\ref{fig:FailureCase}(a), the apparent transformation of Building A is consistent with the indicated camera-motion direction, whereas Building B exhibits a markedly different apparent transformation over the same sequence. This inconsistency reveals the difficulty of maintaining coherent spatial relationships among multiple scene structures under large viewpoint changes. In Fig.~\ref{fig:FailureCase}(b), a small object progressively loses visual integrity and eventually disappears, highlighting the sensitivity of fine-scale content to temporal generation. Fig.~\ref{fig:FailureCase}(c) shows a building undergoing noticeable geometric distortion as the viewpoint rotates, demonstrating the challenge of preserving local structure during rapid rotational motion. In Fig.~\ref{fig:FailureCase}(d), a UAV body that is absent from the preceding visual context emerges in the generated sequence, representing a spurious carrier-related visual artifact. Together, these cases show that MotionScape can expose complementary limitations in global structural consistency, small-object preservation, geometric stability under viewpoint rotation, and content fidelity in highly dynamic UAV-view continuation.

\section*{Data Availability}

The MotionScape benchmark is available on Zenodo~\cite{guo2026motionscape} at \url{https://doi.org/10.5281/zenodo.21954342}, including source metadata, clip-level annotations, and motion-stratification assignments.

\section*{Code Availability}

Code for benchmark reconstruction, motion stratification, and evaluation is publicly available at \url{https://github.com/Thelegendzz/MotionScape}.\ The repository also includes the exact prompts for CLIP-assisted inspection, semantic annotation, and benchmark evaluation.

\vspace{-0.25\baselineskip}
\section*{Funding}
This work was supported by the Key Program of Chinese Academy of Sciences under Grants KGFZD-145-25-38 and RCJJ-145-24-13, and by the Science and Disruptive Technology Program under Grant AIRCAS2024-AIRCAS-SDTP-03.

\vspace{-0.25\baselineskip}
\printbibliography

\end{document}